\pdfoutput=1
\documentclass{article}

\PassOptionsToPackage{numbers}{natbib}
\usepackage[dblblindworkshop, final]{neurips_2026}

\usepackage[T1]{fontenc}
\usepackage[utf8]{inputenc}
\usepackage{amsmath,amssymb}
\usepackage{booktabs}
\usepackage{graphicx}
\usepackage{microtype}
\usepackage{xcolor}
\usepackage{algorithm}
\usepackage[noend]{algpseudocode}
\usepackage[colorlinks=true,allcolors=blue!55!black]{hyperref}
\hypersetup{pdftitle={Metrics That Write Themselves: Evolving an Evaluator from Its Own Blind Spots},pdfauthor={Xing Zhang, Yanwei Cui, Guanghui Wang, Zhihao Lin, Peiyang He}}
\usepackage[capitalise,noabbrev]{cleveref}
\usepackage{caption}


\newcommand{\op}[1]{\texttt{\small #1}}
\newcommand{\dlt}{\ensuremath{\Delta}}
\newcommand{\ntp}{n_{\mathrm{TP}}}
\newcommand{\nfp}{n_{\mathrm{FP}}}

\title{Metrics That Write Themselves:\\
Evolving an Evaluator from Its Own Blind Spots}
\workshoptitle{TAE (Trust-AI-Eval): Can We Trust AI Evaluation?}

\author{%
  Xing Zhang \quad Yanwei Cui \quad Guanghui Wang \quad Zhihao Lin \quad
  Peiyang He\thanks{Corresponding author: \texttt{peiyan@amazon.com}} \\[3pt]
  \normalfont AWS Generative AI Innovation Center
}

\begin{document}
\maketitle

\begin{abstract}
Agents improve quickly against a reliable automatic metric and stall without one, and the applications
that need them most, report generation among them, are the ones nobody knows how to score. Can the
metric write itself? Saying what makes an answer good is hard; pointing at something wrong with one is
easier, so the metric we evolve is a pool of small Python \emph{operators} that each flag a candidate
for one named defect, or abstain, and vote. Asking a model for operators directly
does not work: 183 candidates realise only 96 distinct behaviours, from one narrow region of an enormous
space. EvalCEGAR instead borrows counterexample-guided abstraction refinement from program verification.
It reads the pool as an abstraction and searches for a \emph{collision}, two answers the operators score
identically, one correct and one not. That pair, not a prompt, is the authoring request,
and when a collision defeats every attempt the loop widens what an operator may read rather than
resampling. On MBPP+ and HumanEval+, whose hidden unit tests give exact ground truth, six of eight
runs admit an operator and all six help on 428 unseen tasks, 4 of them significantly, at a median
$+0.0029$ and a best $+0.0065$: 55 lines of Python closing $15.4\%$ of the gap between flagging
nothing and a perfect filter, at a quarter of our best hand-written operator's flags. On the benchmark
the loop never saw, the best of them matches that operator's effect exactly at a third of its flags. Our
15 hand-written operators applied together as one filter \emph{lose} accuracy. An LLM judge on the same information ties that delta on a nearly disjoint
set of candidates, and charges a model call per candidate forever where the operator charges none.
\end{abstract}

\section{Introduction}

A self-improving system needs a metric before it needs a policy. Given a reliable automatic metric,
LLM agents improve rapidly against it: self-refinement, self-debugging and reflective loops all assume
a signal separating better answers from
worse~\citep{madaan2023selfrefine,shinn2023reflexion,chen2024selfdebug}. Without one, progress stalls,
and the substitutes are weak: reference overlap misses semantics by
construction~\citep{papineni2002bleu,lin2004rouge}, and an LLM judge carries position, verbosity and
self-preference biases that belong to the judge, not the
answer~\citep{zheng2023judging,wang2023unfair}. This bites hardest on open-ended output, a report or a
plan, where there is neither a reference answer nor a settled account of what a good one looks like.

This paper asks whether the \emph{metric} can be evolved automatically. What we evolve is not a score
but a pool of \emph{operators}: small Python functions that each read a task and one candidate answer
and either \emph{flag} it for one specific defect, pass it as \emph{clean}, or \emph{abstain}. A pool
becomes a metric by voting: a candidate is rejected when enough members flag it. This form follows an
asymmetry that survives the absence of a reference: saying how good an answer is needs a standard nobody
has written down; pointing at something wrong with it usually does not. Each operator
states one such objection, so it can be read, run and falsified on its own, and the pool's validity
is a measured property rather than an assumed one: every member had to earn its place on a downstream
decision before it joined.

\paragraph{What our first runs showed.}
We first ran the obvious loop: ask the model for an operator, keep it if a gate accepts it, repeat. It
stalls in two ways, and we take both as the problem statement. \emph{Obstacle 1: the
space of operators is enormous, and a model asked for ``an evaluation operator'' samples a tiny, fixed
region of it.} Resampling does not move that region: scoring all 295 operators the study
authored at that narrow interface yields no discovery at all (\cref{sec:ablate}). \emph{Obstacle 2: what
it does write repeats itself behaviourally, and most of it is rejected.} Five runs produced 183 operators that
flag anything, realising only 96 distinct flag sets (\cref{sec:limits}). A rejection teaches nothing either: the loop learns that a candidate failed, not
which distinction it failed to draw.

\paragraph{Our approach.}
We borrow a discipline from program verification: counterexample-guided abstraction refinement
sharpens an abstraction only when a concrete counterexample proves the current one too
coarse~\citep{clarke2000cegar}. The operator pool is that abstraction, mapping each candidate answer to
a \emph{signature}, the vector of verdicts the pool returns for it. Two answers with the same signature
are indistinguishable to the metric, so a pair sharing one while disagreeing on ground truth proves the
metric cannot express the distinction. EvalCEGAR hands that pair to the model as the specification.
Each admission re-partitions the signatures, so the next specification differs by construction: the
target, not the sampler, supplies the diversity pressure (Obstacle 2). When a pair defeats every attempt, the
loop widens the \emph{interface} an operator may read rather than sampling again, since a distinction
the interface cannot see is unreachable at any sample count (Obstacle 1). An operator is admitted only
if it improves the decision the metric is deployed to make, not if it covers more known faults, and
admitted operators vote.

\paragraph{Why we validate on code.}
The motivation is domains \emph{without} a usable metric, yet every number here comes from a domain
that has an exact one: Python problems from MBPP+ and
HumanEval+~\citep{austin2021mbpp,chen2021codex,liu2023evalplus}, whose hidden unit-test suites settle
correctness outright. A method that writes metrics cannot be evaluated where the truth is unknown, so
we build it where the truth is known and withheld. The
operators are domain-specific by construction, so a new domain inherits the loop, not this pool. What transfers is narrow but checkable: the operators never read the oracle, four screens
enforce that they do not reconstruct one (\cref{app:screens}), and oracle labels enter only on the
training split, at a price a human labeller can pay: of our 75 training tasks only 10 carry both a correct
and an incorrect candidate that survive the visible checks, and that was enough to admit operators that
transfer (\cref{app:budget}).

\paragraph{Contributions.}
\begin{itemize}\itemsep1pt \parskip0pt \topsep2pt
\item \textbf{EvalCEGAR.} A loop that evolves an evaluation metric from its own blind spots, the
authoring request being a counterexample it finds for itself (\cref{sec:method}).
\item \textbf{Operators that hold up out of sample.} Six of eight runs admit an operator and all six
help on 428 unseen tasks; the best is 55 lines of Python and matches our best hand-written operator on a
second benchmark at a third of its flags, where a model judge on the same information ties it while
flagging a near-disjoint set (\cref{sec:payoff}).
\item \textbf{What each mechanism is worth.} Ablated inside the loop on identical information: the
admission question alone moves admissions from zero to one, and the narrow interface admits nothing
in 336 attempts (\cref{sec:ablate}).
\item \textbf{Composition solved exactly, not searched.} Scoring all 2.3M subsets the loop can compose
ranks the objective it used to combine them as low as the 5th percentile of that space, and a
class-weighted one, chosen on that same space, at the 91st or above (\cref{sec:selection}).
\end{itemize}

\section{Related work}
\label{sec:related}

\paragraph{Automatic evaluation of open-ended output.} Both substitutes above have been sharpened
without changing shape: learned metrics replace overlap counting with a model's own similarity
judgement~\citep{zhang2020bertscore}, and judge frameworks hand the judge explicit criteria to fill
in~\citep{liu2023geval}, refined by decomposing coarse ones and filtering the
redundant~\citep{shen2026rrd}. The output is still a single scalar whose parts a reader cannot separate.
EvalCEGAR produces a different object, a pool of small, executable, individually falsifiable
\emph{objections}. A pool can gain or
lose one without retraining, which is what makes a metric of this form evolvable. We compare against
both kinds: the judge ties us on the endpoint while dissenting on which candidates to drop
(\cref{sec:payoff}), so the claim is about how a checkable metric is produced, not about beating a
scalar.

\paragraph{LLM-driven program search and CEGAR.} FunSearch evolves programs against a \emph{given}
evaluator~\citep{romera2024funsearch}, and automated agent design searches agent code against a
\emph{given} benchmark~\citep{hu2024adas}. Ours is the complement: with the evaluator itself as the
artefact there is no fixed fitness function, so the design problem moves from how candidates are mutated
to what a candidate must prove before it is kept. Closest in output are methods that induce executable checks
with an LLM and combine them, either into a compact set of Python verifiers whose joint satisfaction
approximates a labelled objective~\citep{pezeshkpour2026autopyverifier} or into an unweighted vote over
model-written yes/no questions~\citep{griffin2025rrf}. Those search for checks that predict a label; we search for the
counterexample showing the current ones cannot, and admit on the deployed decision instead. We run that
recipe as a baseline (\cref{sec:ablate}).
CEGAR is a verification technique~\citep{clarke2000cegar}; to our knowledge its transfer to
\emph{metric} synthesis is new (\cref{sec:method}).

\paragraph{Diversity and test generation.} Quality-diversity search keeps a population from collapsing by
maintaining an explicit archive of behaviours~\citep{mouret2015mapelites}; our diversity pressure comes
instead from re-partitioning the target after each admission, and \cref{sec:limits} measures what that
does not buy. Nor is this test generation: a property-based tester needs a property to
falsify~\citep{claessen2000quickcheck} and differential testing needs a second
implementation~\citep{mckeeman1998differential}, and our setting supplies neither.

\section{Preliminaries}
\label{sec:setup}

\paragraph{Tasks and data.}
Every task is a Python programming problem from MBPP+ or HumanEval+ as distributed by
EvalPlus~\citep{liu2023evalplus}, whose hidden test suite is the exact oracle. It labels the training
samples the loop targets and admits on, and nothing but the scoring script reads it anywhere else. A task statement also shows a few example calls with their expected
results, which the solver sees too; we call these the \emph{visible checks}, and they are not held-out
information. We drew 150 MBPP+ tasks, split 75 to run the loop on and 75 held out, and added the 228
remaining MBPP+ tasks and 125 of the 164 HumanEval+ tasks to the held-out
side.\footnote{HumanEval+ statements contain no asserts, so its visible checks are synthesised from the docstring
examples; we keep the 125 tasks that yield at least one such check the reference solution passes.}
Results are reported on those 428 unseen tasks as \emph{MBPP+ held-out} (75), \emph{MBPP+ additional}
(228) and \emph{HumanEval+} (125), the only group from a different task distribution.

\paragraph{Selection accuracy.}
For each task $t$ we sample many model solutions and let $V(t)$ be those that pass the visible checks,
duplicates included: the candidates a deployed system would plausibly accept, so the faults that matter
are the ones surviving them. A metric must choose among them. It drops from $V(t)$ what it flags,
leaving the kept set $K(t)$, and one sample is drawn uniformly from $K(t)$, or from $V(t)$ if the metric
flagged everything. \emph{Selection accuracy} is the probability that this sample is correct, averaged
over tasks; $\dlt$ always means its change against the metric that flags nothing.

\paragraph{Decidability and headroom.}
A task is \emph{decidable} only if $V(t)$ holds both a correct and an incorrect sample; on any other
task the drawn sample's correctness is settled whatever the metric does. Only 58 of the 428 held-out
tasks are decidable at our sampling depth, so selection accuracy over all 428 runs from $0.7170$, when
nothing is flagged, up to $0.7593$ for a perfect operator. Absolute $\dlt$ values are small for that
reason, so we also report each as a fraction of that reachable range, the \emph{headroom}.
The endpoint is also \emph{recall-dominated}: a flagged correct sample merely leaves the kept set, while
a caught incorrect one changes what remains to draw from.

\paragraph{What we compare against.}
Before building the loop we wrote 15 operators by hand, ten static checks on the candidate's syntax tree
and five that run it. They are the pool the loop starts from and, applied together as one filter, our human baseline. Our
strongest single hand-written operator, called the \emph{comparator} throughout, flags a candidate whose
behaviour disagrees with the plurality of the other samples for its task. The second baseline is an
LLM judge, asked one question per candidate given exactly a level-1 operator's information, in four
configurations over two models and two prompts (\cref{app:judge}).

\paragraph{The model that writes operators.}
One frozen model (Claude Opus 4.7, no
fine-tuning and no gradient update anywhere in this work) emits operator source code, at 47--48 calls
per run under a 60-call ceiling across 8 rounds. The faulty candidates its operators are judged on come from
weaker models (Llama 3.1 8B Instruct and
Claude Haiku 4.5, with Amazon Nova Micro and Mistral 7B Instruct added for the deeper draw of
\cref{sec:limits}), so the model that writes bad code never wrote the operators that judge it. What EvalCEGAR ships, the
operator pool and its vote, is pure Python with \emph{no} inference-time model call.

\paragraph{Keeping the loop honest.}
Four screens reject an authored operator before the gate sees it, each closing a way an earlier round of
this work reconstructed the oracle by accident: deciding from the candidate's syntax without running it,
keying on a constant the fault generator happens to emit, branching on prompt keywords, and reaching a
flag by text match while executing the candidate elsewhere. None fires on any of the 15 hand-written
operators, so they are not a ban on ordinary detectors (\cref{app:screens}). Every $\dlt$ is tested
against a size-matched shuffle null: per task the operator's own flag count is redrawn from that task's
own samples, so the null fixes how much it flags and randomises only where the flags land. That licenses
``this rule beats flagging nothing'' but not ``rule $A$ beats rule $B$'', the claim \cref{sec:limits}
reports on, and multiple comparisons carry a Bonferroni correction within a family of related tests and a
Benjamini--Hochberg one across a screen of many~\citep{benjamini1995controlling}. And \emph{registered}
means the prediction was written down before the script that tests it existed.

\section{Method}
\label{sec:method}

\begin{figure}[t]
\centering
\includegraphics[width=\linewidth]{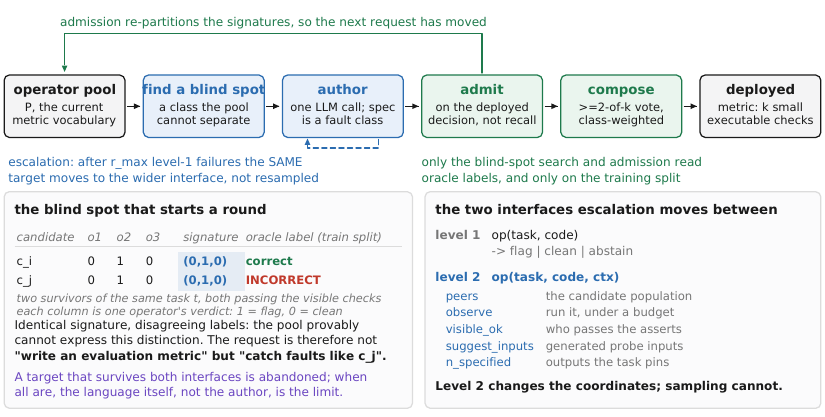}
\caption{EvalCEGAR. The authoring request is a \emph{blind spot}: answers the pool maps to one
signature while ground truth disagrees (left). Admission re-partitions the signatures, so the next
request has moved, and a blind spot that resists level 1 escalates (right) rather than being resampled.}
\label{fig:loop}
\end{figure}

\paragraph{In one sentence.}
Find two candidate answers the \emph{current} pool scores identically, one correct and one incorrect;
ask the model for an operator that catches faults of that kind; widen what an operator may read when
the request proves unreachable; and admit on whether the deployed decision improves. \Cref{fig:loop}
is the loop and \cref{alg:loop} one round of it.

\begin{algorithm}[t]
\small
\caption{EvalCEGAR, one round on one target. $\dlt_D$ is the change in selection accuracy on the
training tasks $D$. Ground-truth labels $y$ are read on the training split only, at lines 2 and 10;
\textsc{Author} sees the specification and the interface, never $y$.}
\label{alg:loop}
\begin{algorithmic}[1]
\Require operator pool $P$, training tasks $D$ with candidate sets $V(t)$ and labels $y$, level-1
retry budget $r_{\max}$, total attempt budget $R$
\State $\sigma_P(c) \gets \big(o(t,c)\big)_{o \in P}$ for every $c \in V(t)$, $t \in D$
  \Comment{the pool \emph{is} the abstraction}
\State $\mathcal{C} \gets \{\,$signature classes containing both a $y{=}1$ and a $y{=}0$ candidate$\,\}$
  \Comment{the blind spots}
\If{$\mathcal{C} = \emptyset$} \Return $P$ \Comment{the pool already separates every labelled pair}
\EndIf
\State $\kappa \gets \arg\max_{\text{class} \in \mathcal{C}} |\text{class}|$ \Comment{the target}
\State $\mathrm{spec} \gets$ every incorrect training sample that fails as $\kappa$'s do, and every
  correct one \Comment{not just the pair}
\State $\ell \gets 1$ \Comment{interface level: \op{op(task, code)}}
\For{$r = 1 \dots R$}
  \State $o \gets \textsc{Author}(\mathrm{spec}, \ell)$ \Comment{one LLM call; a stated request, not ``write a metric''}
  \If{$o$ fails a leakage screen, does not shrink $\kappa$, or is a function of $\sigma_P$}
    \textbf{continue} \EndIf
  \If{$\dlt_D(P \cup \{o\}) > 0 \;\wedge\; \mathrm{helped} \geq 3 \;\wedge\; \mathrm{helped} > \mathrm{hurt}$}
    \State \Return $P \cup \{o\}$
      \Comment{admission re-partitions $\sigma_P$, so the next $\kappa$ has moved}
  \EndIf
  \If{$r = r_{\max}$} \State $\ell \gets 2$
    \Comment{\emph{same} $\kappa$, wider interface \op{op(task, code, ctx)}}
  \EndIf
\EndFor
\State \Return $P$ \Comment{$\kappa$ abandoned: unreachable at either interface}
\end{algorithmic}
\end{algorithm}

\paragraph{Collision targeting.}
A \emph{collision}, or blind spot, is a signature class holding both correct and incorrect
candidates, and the largest one in the pool becomes the next target. The colliding pair locates the
request but does not state it: the request covers every training sample that fails in the same way,
because a specification naming two samples is satisfiable by a lookup table
(\cref{app:screens}). A target that survives every attempt at both interfaces is abandoned,
and once every target has been abandoned the loop returns an \emph{insufficiency certificate} for the
operator vocabulary: the metric \emph{language}, not the author, is the limit. That is what makes a failed
run informative: it names the distinction the vocabulary could not express.

\paragraph{Interface escalation.}
Level 1 is \op{op(task, code)}. When $r_{\max}{=}3$ level-1 attempts fail on a target, the \emph{same}
target escalates to level 2, \op{op(task, code, ctx)}, where \op{ctx} lets the operator see the
candidate's \emph{peers} and run any of them on inputs of its own choosing, under a hard budget of 600
observations per operator call over at most 16 peers (\cref{fig:loop}, right, and \cref{app:screens}). This changes the coordinates of the search instead of resampling
within them: if the useful operator is not expressible at level 1, no amount of sampling finds it.

\paragraph{Admission on the deployed decision.}
The final gate is a floor on the \emph{decision}, evaluated on the training split: $\dlt > 0$, at
least 3 tasks helped, helped $>$ hurt. It is \emph{not} a floor on fault-class recall, which matters
more than it sounds: a sparse, precise operator fails a recall floor by construction and is exactly
what the endpoint wants. The recall-gated arm rejected our best operator for covering none of the
140 known faults the pool still missed (\cref{app:ablate}).

\paragraph{Composition by error independence.}
Admitted operators are composed by a vote: a candidate is flagged when at least $m$ of the $k$
operators in a chosen subset $S$ flag it, with $m{=}2$ and $k = |S| \leq K_{\max}{=}6$ unless stated
otherwise. The subset is the one maximising an objective computed on the labelled training samples
alone, with no development split, no tuned threshold and no tie-break, and unit weights rather than
learned ones, since a unit-weight vote is hard to beat when the positive class is
scarce~\citep{griffin2025rrf}. The objective was originally
$J = \mathrm{TP} - \mathrm{FP}$ over \emph{corroborated} flags, those on which the vote fires;
\cref{sec:selection} shows $J$ works \emph{against} deployment and replaces it with
$\mathrm{TP} - \lambda\,\mathrm{FP}$ at $\lambda = \ntp/\nfp$ read off the training split.

\section{Results and analysis}

Three questions: whether the metric works on unseen tasks (\cref{sec:payoff}), what each mechanism
contributes when it is removed (\cref{sec:ablate}), and how good the composition step is once its whole
search space can be scored exactly (\cref{sec:selection}).

\subsection{Main result}
\label{sec:payoff}

\begin{table}[t]
\caption{The authored operator against the strongest hand-written one and an LLM judge, on the 428
held-out tasks and their 2592 samples. \% of headroom is $\dlt$ as a fraction of the $0.0424$ a
perfect operator could add (\cref{sec:setup}); \emph{calls} is model requests per candidate scored.
$^{*}$best of a family picked by held-out $\dlt$, four judge configurations and six admissions, so
uncorrected for the pick; $0.004$ and $0.006$ under Bonferroni.}
\label{tab:headline}
\centering\small
\setlength{\tabcolsep}{4.4pt}
\begin{tabular}{@{}lrrrrr@{}}
\toprule
& $\dlt$ & \% of headroom & $p$ & flags & calls \\
\midrule
hand-written comparator                      & $+0.0069$ & $16.3$ & $0.0130$ & 515 & 0 \\
all 15 hand-written operators as one filter  & $-0.0067$ & $-15.8$ & n/a & 1425 & 0 \\
LLM judge, best of four configurations       & $+0.0070$ & $16.5$ & $0.0010^{*}$ & 158 & 1 \\
\textbf{loop-authored operator}              & $+0.0065$ & $15.4$ & $0.0010^{*}$ & 126 & 0 \\
\bottomrule
\end{tabular}
\end{table}

\Cref{tab:headline} compares single operators. The loop's is 55 lines of Python and reaches
$94.2\%$ of the comparator's effect with a quarter of its flags. Its $+0.0065$ rises to $+0.0481$
over the 58 decidable tasks alone (\cref{tab:transfer}), and its flags overlap the closest hand-written
operator at Jaccard $0.245$, against the ${>}0.84$ of this study's own rediscoveries: a shared idea
drawn differently, not a copy. What it does reads in one sentence: keep
the peers that pass the prompt-visible asserts, run them and the candidate on perturbed inputs, and flag
disagreement with the surviving plurality. Every admitted level-2 operator is a variant of that
algorithm, since the loop rewards a new distinction rather than a new way to draw one (\cref{sec:limits}). The same table holds the contrast that
motivates the loop: in a domain people find \emph{easy}, the 15 hand-written operators as one
filter \emph{lose} accuracy.

\paragraph{A model judge on the same information.}
Asked one question per candidate under a level-1 operator's exact interface, and scored as a metric on
the identical endpoint and null, the best of four judge configurations reaches $+0.0070$
(\cref{tab:headline}), so we do not claim to beat it. The two flag sets meet at Jaccard $0.105$, 27 flags of 257; each side's
exclusive flags are null alone ($+0.0010$ at $p{=}0.068$, and $-0.0007$); and the 27 they agree on are
worth $+13/{-}1$ tasks, with the union above both parts at $+0.0073$. A checkable operator and a model
judgement fail on different inputs, so their agreement is a high-precision region neither reaches
alone. Cost separates them permanently: 47 authoring calls and nothing per candidate, against one call
per candidate forever, 1762 for a single pass here. Two asymmetries belong with that comparison: the author model judges at $+0.0040$, below what it authors, and both
benchmarks are public, so a judge may have memorised solutions where an operator cannot
(\cref{app:judge}).

\paragraph{A pool, not one operator.}
Eight runs differ only in the model's sampling, so their spread is EvalCEGAR's own sampling
distribution: 6 of 8 admit an operator, all 6 are
positive out of sample (median $+0.0029$, range $+0.0009$ to $+0.0065$, 4 of 6 individually
significant), all 6 are at level 2, and all four fitting screens are clean on all 6.
\Cref{tab:headline}'s row is that range's top, picked on held-out $\dlt$, and the distribution rather
than that row is the result. Across the whole
study the gate has admitted 26 operators: on the held-out tasks 25 are non-harmful and 17 are individually significant against their own
nulls, where chance predicts $1.3$, at a median of $+0.0036$. The 15 hand-written operators are
not such a pool: 9 flag anything at all, 5 are non-harmful, and 1 is significant. Admission reads the
training split alone, so this is transfer.

\paragraph{Transfer to an unseen benchmark.}
Two facts stand out (\cref{tab:transfer}). Of 13 rules measured on all three pools, 12
loop-authored operators and the comparator, this one alone is positive \emph{and} clears its own
per-pool null on all three, where the comparator reaches $p<0.05$ on HumanEval+ alone. And on
HumanEval+, the pool from a different task distribution, it matches the comparator exactly and beats it on
parsimony: the same 10 helped tasks, the same single hurt task and the same $+0.0125$, which is $28.8\%$
of that pool's headroom, with 36 flags against 103. Each pool was first checked to hold enough decidable tasks for a
perfect operator to register, so a zero would mean absence, not no power. Transfer also holds along a
second axis, to 9320 candidates from three generators the composed rule never saw, at $22.1\%$ of that
draw's headroom against $18.9\%$ here (\cref{app:peers}).

\begin{table}[t]
\caption{Per-pool transfer of the authored operator, each against its own shuffle null ($b{=}1000$
resamples). HumanEval+ differs from the authoring pool in oracle module, prompt style and difficulty,
and nothing was re-tuned. $+/-$ counts tasks helped and hurt; false alarms is the rate at which
\emph{correct} samples are flagged, authored / hand-written. Per-pool headroom: $0.0392$, $0.0429$,
$0.0433$.}
\label{tab:transfer}
\centering\small
\setlength{\tabcolsep}{4.4pt}
\begin{tabular}{@{}lrrrrrrr@{}}
\toprule
pool & tasks & decidable & $\dlt$ & $p$ & $z$ & $+/-$ & false alarms \\
\midrule
MBPP+ held-out              &  75 &  6 & $+0.0047$ & $0.0400$ & $1.74$ & $+2/{-}1$ & $0.054$ / $0.168$ \\
MBPP+ additional            & 228 & 34 & $+0.0038$ & $0.0020$ & $3.38$ & $+8/{-}4$ & $0.041$ / $0.129$ \\
\textbf{HumanEval+}         & 125 & 18 & $\mathbf{+0.0125}$ & $\mathbf{0.0010}$ & $\mathbf{6.28}$ & $+10/{-}1$ & $0.027$ / $0.071$ \\
\bottomrule
\end{tabular}
\end{table}

\subsection{Mechanism ablations}
\label{sec:ablate}

\begin{table}[t]
\caption{The two mechanism ablations, each run inside the loop on its own candidates, over the same three
seeds and the same training-split labels, with only the named component changed. \emph{Calls} is the
model's authoring calls per seed, so neither ablation is a starved run.}
\label{tab:ablate}
\centering\small
\setlength{\tabcolsep}{8pt}
\begin{tabular}{@{}llcc@{}}
\toprule
arm & what changes & calls & admissions \\
\midrule
the loop (reference)        & --                      & 47 / 48 / 47 & \textbf{1 / 1 / 1} \\
recall floor at the gate    & the admission objective & 48 / 48 / 48 & \textbf{0 / 0 / 0} \\
level 1 only, no escalation & the interface           & 48 / 48 / 48 & \textbf{0 / 0 / 0} \\
\bottomrule
\end{tabular}
\end{table}

Removing either mechanism stops admission outright (\cref{tab:ablate}): each is necessary, not merely
helpful. The narrow-only arm's failure is not a budget artefact, because level 1's own
ceiling is below chance: scoring all 295 level-1 operators the study authored on the pooled 503 tasks,
with \emph{no split and no selection cost}, gives 9 at $p<0.05$ against 14.75 expected by chance, so
Benjamini--Hochberg returns \emph{no discoveries} at a 10\% false-discovery rate.

An independent search agrees on where that wall is. Commissioning checks against a balanced labelled
sample and admitting them for \emph{predicting the label}, as the induction methods of
\cref{sec:related} do, admits 2 of 48 candidates at level 1, whose $+0.0016$ does not clear its own null, and
6 of 16 at level 2 for $+0.0054$ under a training-only combiner (\cref{app:judge}). Two searches sharing nothing but the
interface both fail at level 1 and both work at level 2, which also bounds what the recall-floor row
claims: an admission objective must not reward coverage, but predictive fit does not fail the way recall
gating does.

\paragraph{The gate ledger.}
Recording \emph{every} gate consultation locates that wall exactly.
Admission needs three helped
tasks; the narrow interface reaches two and stops there 55 times with no task hurt, and none of its 124
consultations reaches three, while the wider interface reaches three 6 times and every one was admitted
(\cref{fig:wall}). The extension does not buy the \emph{direction}, which the narrow interface finds
easily; it buys \emph{generality}, the third distinct task. Targeting itself buys
candidate viability rather than admissions: an operator abstains everywhere 34.5\% of the time undirected
against 2.1\% directed (\cref{app:cert}).

\subsection{Composition, enumerated}
\label{sec:selection}

\begin{figure}[t]
\centering
\includegraphics[width=\linewidth]{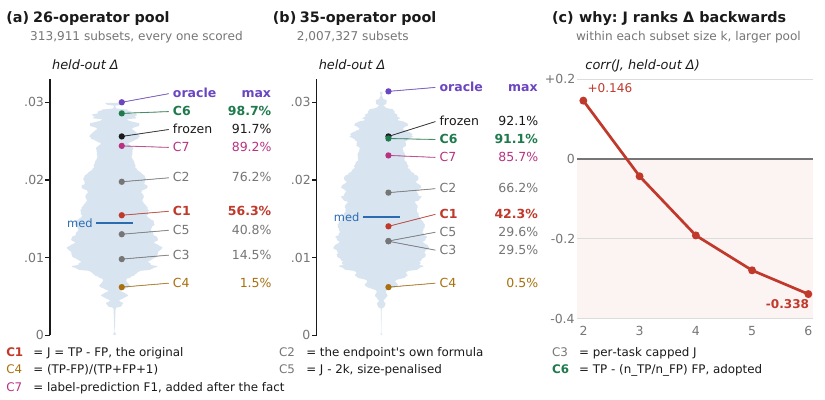}
\caption{The composition search space, enumerated. \textbf{(a,b)} Held-out $\dlt$ of every
$\geq2$-of-$k$ subset ($k \leq 6$) of the 26- and 35-operator pools, as a density, with C1--C7 of
\cref{tab:criteria}, the frozen rule and the oracle argmax each at its exact percentile. C7 was added after the fact.
\textbf{(c)} Spearman correlation between $J$ and the endpoint \emph{within} each subset size on the
larger pool: positive only at $k{=}2$, so the misalignment is no size artefact.}
\label{fig:land}
\end{figure}

Composition chooses a subset of size at most 6 from the loop's admitted operators, with no
hand-written operator in either pool. Both pools collect the gate's admissions across the study's
authoring arms and seeds, 26 operators and 35, the larger containing the smaller, and they
realise 17 and 20 distinct flag behaviours, giving 313{,}911 and 2{,}007{,}327 subsets, small enough to
\emph{enumerate}. We scored every one
on the held-out endpoint through the same code used everywhere else (\cref{fig:land}), so
questions previously answered by inference become arithmetic and no $p$-value appears in this
subsection. Two terms recur: a subset is \emph{vacuous} when its vote flags nothing, and the
\emph{oracle argmax} is the best held-out $\dlt$ in the enumeration, an upper bound no training-data
rule should reach. Every result spans four \emph{cells}: each pool at each of two
pinned settings of the peer set a level-2 operator reads.

\paragraph{The defect.}
$J$ ranks its own argmax anywhere from the 4.9th to the 85.7th percentile of the space it searches
across the four cells (C1 in \cref{tab:criteria}), against an oracle argmax of roughly \emph{twice} its
delta ($+0.0300/{+}0.0314$ against $J$'s $+0.0155/{+}0.0140$), and its within-$k$ correlation with the
endpoint is positive only at $k{=}2$, falling to $-0.338$ by $k{=}6$ (\cref{fig:land}c). Enumeration
also shows the cause is \emph{bias, not variance}: refitting on more labelled data converges
(16--18 distinct selections at $n{=}210$ to 3--5 at $n{=}1195$), and it converges \emph{below} the
\emph{frozen rule}, the subset pinned under an earlier rule of thumb. So freeze the selection rather
than refitting it.

\paragraph{The repair.}
We registered six selection criteria before running, each computed from the training split alone
($\ntp{=}222$ true and $\nfp{=}1183$ false flags), ranked by the unchanged $\geq2$-of-$k$ rule and
scored against the exact landscape they are therefore chosen on (\cref{tab:criteria}). Balanced
accuracy is best of the six in all
four cells, closing 39.8--90.2\% of the gap to the oracle argmax, and it also beats label-prediction
$F_1$, the criterion the induction baseline admits on, in all four.

\paragraph{Why it works.}
Our registered prediction blamed aggregation, that $J$ counts per \emph{sample} while the endpoint
averages per \emph{task}; but C2, the endpoint's own formula, never reaches the 85th percentile, so that
was not the defect. What matters is the class weight:
balanced accuracy is monotone in $\mathrm{TP}-0.188\,\mathrm{FP}$, penalising a false alarm
$5.33\times$ less than $J$'s 1:1, and C4 falls to the 1.5th percentile or below. That is the recall
dominance of \cref{sec:setup} resurfacing as a mis-specified objective, and it means \emph{refitting is
safe once the objective is right}, which a self-evolving library needs. Two registered checks hold: a
split-half protocol on the same tasks re-selects C6 in 4 of 4 cells, and on a 12-point sweep HumanEval+'s optimal
$\lambda$-plateau contains the MBPP training ratio in 4 of 4 cells while $\lambda{=}1$ never does, so
the weight is a property of selection endpoints, not of those labels (\cref{app:robust}).

\paragraph{Safety over the whole closure.}
Of the $2{,}007{,}292$ non-vacuous subsets of the larger pool, \emph{zero} harm the endpoint (minimum
$+0.0007$): admission makes composition safe over its \emph{entire} combinatorial closure, exactly and
with no power argument. The converse holds on the 15 hand-written operators, which admission accepts
\emph{none} of: the same combiner converges there to flagging nothing.

\section{Negative results and limitations}
\label{sec:limits}

\paragraph{Behavioural duplication.}
The pool repeats itself behaviourally rather than textually: no source-text pair among 187 candidates
reaches Jaccard $0.8$, yet the 183 that flag anything realise only 96 distinct flag sets
(\cref{fig:neg}). Neither cure worked. Targeting made it worse, the directed arm being the
\emph{least} diverse of three, and a novelty term at admission recovers a genuinely new operator in only
2 of the 11 candidates it blocks; handed the incumbent it duplicated, the model duplicates it again. So
the restriction sits in the operator prior, not at the gate, which argues for an explicit behavioural
archive~\citep{mouret2015mapelites}.

\paragraph{Waste, not exhaustion.}
Nine of 14 admissions on three fresh seeds rediscovered an operator already selected, and a composition
headline of $+0.0076$ on that pool did not survive the rediscovery screen we then added
(\cref{app:redundancy}). Rarefaction over behavioural classes~\citep{chao1987estimating} still yields 2.00 new
classes at the sixth seed: wasteful, not exhausted.

\paragraph{What the numbers do not show.}
Every $\dlt$ is measured against flagging nothing, never against another rule: no two rules separated in
9 paired tests~\citep{wilcoxon1945individual}, their differences living on 2--9 tasks, and the three
loop operators that appeared to beat the comparator were rediscoveries of it. A level-2 verdict is also
a function of the peer electorate: growing it threefold changes a verdict for all 35 operators, so
their flags need a pinned electorate, though held-out $\dlt$ moves up rather than down
(\cref{app:peers}). What admission discards is unmeasured (\cref{app:cert}). Both benchmarks come from
one distribution family, so an out-of-sample \emph{task list} needs a third oracle port. And the loop is
not the only route to an operator of this quality: a judge ties its delta, so what it buys is a
complementary error profile, not a higher ceiling.

\section{Conclusion}

An evaluation metric can be authored automatically, and the artefact is small enough to read: 55 lines
of Python, no weights, closing $15.4\%$ of the distance to a perfect operator on 428 unseen tasks. The loop asks with a counterexample rather than a prompt, and
three rules behind it carry over to any pool of executable checks:
\begin{itemize}\itemsep1pt \parskip0pt \topsep2pt
\item \textbf{Admit on the decision, not on coverage.} That change alone moved admissions from none to
one per seed.
\item \textbf{When nothing resolves a counterexample, widen the interface.} 336 attempts at the narrow
interface admit none; the wider one admits six.
\item \textbf{On a selection endpoint, weight the classes.} $\mathrm{TP}-(\ntp/\nfp)\mathrm{FP}$, read
off the training labels, reaches the top decile of a space whose precision-flavoured objective sits in
the bottom $2\%$. That ranking is retrospective on this one landscape, not an independent validation.
\end{itemize}
Where unit tests exist, run the unit tests. What remains is the setting that motivated the method,
where an unresolved counterexample is the only evidence that a metric's vocabulary, and so the claim it
can support, has run out.

\bibliographystyle{plain}
\bibliography{refs}

\appendix

\section{Request scope, screens and nulls}
\label{app:screens}

The request handed to the author covers a whole fault kind rather than the two colliding samples,
because a two-sample request is satisfiable by a lookup table: an earlier round of this work measured
exactly that, with operators firing on a handful of training faults and on none of a held-out class. The
two kinds are faults that already fail the visible checks and faults that pass them but fail the hidden
suite; the request covers whichever kind the colliding sample belongs to.

The subtlest of the four screens of \cref{sec:setup} is the prompt-keyword one: an operator that
branches on a keyword answers from the task's identity rather than from the candidate in front of it,
which is oracle reconstruction by dispatch table.

Level 2's \op{ctx} exposes the candidate's peers, a runner, the prompt-visible asserts and a generator
of inputs, and nothing that reads the hidden suite. Inputs come from perturbing the arguments of the
prompt-visible asserts, which the solver also saw, with the number of unperturbed ones exposed so that an
operator can tell the two apart. An operator that exceeds the observation cap abstains rather than
stalling.

The shuffle null of \cref{sec:setup} uses $b{=}1000$ resamples throughout, seeded so that a rerun
reproduces them exactly.

\section{The transfer budget}
\label{app:budget}

Because oracle labels enter only on the training split, the cost of moving EvalCEGAR to an oracle-free
domain is the cost of that split's labels, and it must be quoted in \emph{decidable} examples: an
example informs the loop only if it carries both a good and a bad candidate that survive the visible
checks. Of our 75 training tasks 10 are decidable, giving 80 (correct, wrong) pairs and four
distinguishable levels of the training objective. That is demonstrably enough to \emph{admit} operators
that transfer ($+0.0065$) and demonstrably \emph{not} enough to tune one scalar: a $\geq4$-of-26 vote is
worth $+0.0087$ held-out, and neither training $\dlt$ nor cross-validation inside training can
distinguish it from the union.

Three measurements complete the budget. \emph{(i)} The crossover for tuning aggregation is 20--30
decidable tasks: below it the choice collapses to the union in 9 of 9 directions; above it, rotating a
held-back fold recovers $+0.0077$ ($p\leq0.0004$, identical at $m{=}2,3,4$), 57\% of the oracle's
gain. \emph{(ii)} A better gate removes that requirement rather than paying it: 9 operators from three
seeds of the target-only gate (\cref{app:cert}) score $+0.0071$ ($p{=}0.0022$) as a plain union with no
threshold to set, so gate quality and aggregation are substitutes. \emph{(iii)}
Composition changes the currency altogether, and this is the cheapest line item: error independence
needs \emph{labelled flags}, not decidable tasks, and the same training split holds 426 of them. The
$\geq2$-of-$k$ rule matches \emph{(i)}'s rotated selection with 5 operators instead of 26 and nothing to
tune, at a measured price of $\sim$60 labelled candidates (at half that sample it degrades to
$+0.0065/{+}0.0033$). So the budget line for a new domain is $\sim$10 decidable examples to author and
admit plus $\sim$60 labelled candidates to decide which admitted operators vote, and the second number
is the one a practitioner can actually produce, because it does not require a single example to hold
both a correct and an incorrect candidate.

\section{Targeting versus accepting}
\label{app:cert}

An early ablation on a synthetic bank favoured the collision 4 admissions to 0, with 31 of 36
undirected candidates rejected as \emph{untargeted}. That did not replicate on the natural bank and we
discard the claim it licensed. What did replicate there is the viability effect \cref{sec:ablate}
quotes: on the 422 training witnesses, 49 of 142 undirected candidates return no verdict anywhere,
$34.5\%$, against 4 of 187 directed, $2.1\%$. A later single-flag experiment
settles the division of labour instead:
keeping the collision as the \emph{target} but dropping it as the \emph{acceptance test} raises
distinct-metric yield from $0.38$ to $1.67$ per seed ($4.4\times$) with peak quality unchanged
($+0.0063$ vs $+0.0065$), and moves the binding rejection stage to the objective (60.3\% of
rejections, up from 28.6\%). So the collision is doing two separable jobs, and only the first is worth
its cost. What the gate discards is unmeasured in the other direction too: one operator it turns down
for flagging nothing in training is our most reliable one held out.

\section{The mechanism ablations in full}
\label{app:ablate}

Both arms run inside the loop on the loop's own candidates, so neither is a re-scoring of operators
authored under the other. Both read the same training-split labels, and no gate call in either arm hit
its abort guard. The floor the recall-gated arm applies asks an operator to fire on at least 30\% of
the 140 fault classes the pool still misses, at no more than a 2\% false-alarm rate on the samples that
must pass. The narrow-only arm was also given double the level-1 attempts per target, 48 per
seed against the full loop's 24, and pooling its 144 with every level-1 attempt the eight full runs made
gives the 336.

\section{The gate ledger in full}

\begin{figure}[h]
\centering
\includegraphics[width=\linewidth]{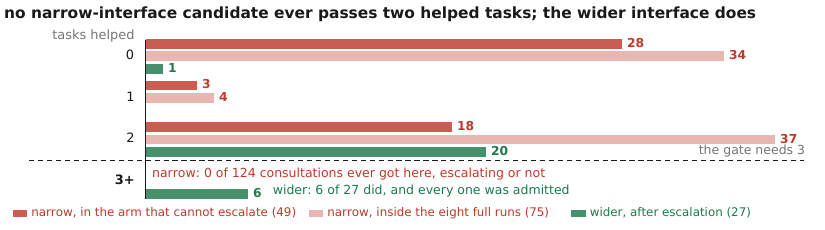}
\caption{Every consultation of the admission gate, not just the ones that ended in an admission, by
how many tasks the candidate helped on the training split the gate reads. Admission needs three, and
only the wider interface ever gets there, whether or not the narrow run could escalate.}
\label{fig:wall}
\end{figure}

Of the 336 level-1 attempts of \cref{tab:ablate}, the
124 that survived every earlier screen are the ones the gate saw and \cref{fig:wall} draws; a target is
attempted at the narrow interface first, so most of an escalating run's consultations are narrow ones and
reading a run's ledger as one interface would over-attribute them. Each of the six operators admitted
across the eight runs has a recorded level-1 rejection immediately before it. The ledger doubles as the
sensitivity analysis for the admission floor: at helped $\geq2$ the narrow interface would admit 55 of
its 124 consultations, so the floor is what an interface has to clear, and what those admissions would
transfer to is unmeasured.

\section{The selection criterion: registration, robustness, and the class weight}
\label{app:robust}

All six criteria, their formulas and their predicted ranking were written down before the landscape was
scored, so \cref{tab:criteria} reports where each choice lands rather than which choice was searched for.

\begin{table}[ht]
\caption{Six a-priori criteria, and one added afterwards, against the enumerated landscape of
\cref{fig:land}. Cells run (26-operator pool, 35-operator pool) at each peer-set setting. C6 is the
rule we adopt, and nothing is tuned: $0.1877$ is the training split's class ratio $\ntp/\nfp$. C7 is
the induction baseline's own admission signal and the only row not registered in advance.}
\label{tab:criteria}
\centering\small
\setlength{\tabcolsep}{4.4pt}
\begin{tabular}{@{}llcc@{}}
\toprule
& criterion (training split only) & landscape percentile, 4 cells & held-out $\dlt$ \\
\midrule
C1 & $J = \mathrm{TP}-\mathrm{FP}$ \; (the original) & 56.3 / 42.3 / 85.7 / \phantom{0}4.9 & $+0.0079$ \dots $+0.0220$ \\
C2 & the endpoint's own formula on training data & 76.2 / 66.2 / 78.3 / 82.1 & $+0.0184$ \dots $+0.0209$ \\
C3 & per-task capped $J$ & 14.5 / 29.5 / 50.4 / 44.1 & $+0.0098$ \dots $+0.0138$ \\
C4 & precision-flavoured $(\mathrm{TP}{-}\mathrm{FP})/(\mathrm{TP}{+}\mathrm{FP}{+}1)$ & \phantom{0}1.5 / \phantom{0}0.5 / \phantom{0}0.4 / \phantom{0}0.1 & $+0.0043$ \dots $+0.0062$ \\
C5 & size-penalised $J - 2k$ & 40.8 / 29.6 / 32.5 / \phantom{0}2.3 & $+0.0072$ \dots $+0.0130$ \\
\textbf{C6} & \textbf{balanced accuracy} $\equiv \mathrm{TP} - (\ntp/\nfp)\,\mathrm{FP}$ & \textbf{98.7 / 91.1 / 92.7 / 94.1} & $\mathbf{+0.0236}$ \dots $\mathbf{+0.0286}$ \\
C7 & label-prediction $F_1$ \; (not registered) & 89.2 / 85.7 / 85.0 / 85.8 & $+0.0218$ \dots $+0.0244$ \\
\bottomrule
\end{tabular}
\end{table}

Both registered robustness checks behind \cref{sec:selection} are reported here in full.
\emph{(i) Split-half.} Choosing the criterion on half A of the held-out tasks and reporting on half B,
with the split fixed in the registration, re-selects C6 in 4 of 4 cells at the 85.2--98.2nd percentile
of half B's own landscape, against $J$'s 9.3--82.4th.
\emph{(ii) Per-benchmark.} The endpoint
decomposes exactly by task, so the same held-out tasks
re-partition by provenance at zero cost, recombining to the reported landscape maximum to 6 decimal
places. On HumanEval+, which shares only the operator pool with the training split, C6 ranks at the
95.9 / 97.6 / 95.9 / 97.5th percentile, its best pool of the three, beating $J$ in 4 of 4 cells and
separating from it in a paired test in 3 of 4. Reading the same 428 tasks three ways adds no independent
evidence for the overall comparison, but it does rule out the gain being an artefact of the benchmark
the labels came from.

\begin{figure}[h]
\centering
\includegraphics[width=\linewidth]{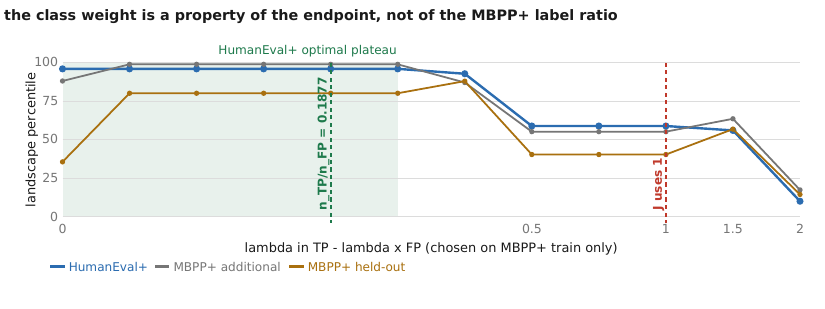}
\caption{Sweeping $\lambda$ in $\mathrm{TP}-\lambda\,\mathrm{FP}$, one line per held-out pool:
HumanEval+'s optimal plateau (shaded) contains $\lambda{=}\ntp/\nfp{=}0.1877$, read off the MBPP+
training split, and excludes $J$'s $\lambda{=}1$.}
\label{fig:sweep}
\end{figure}

\paragraph{The sweep (\cref{fig:sweep}).} $\lambda = \ntp/\nfp$ is read off the MBPP training split, so
the question is whether it is a property of those labels or of selection endpoints generally. Twelve
values of $\lambda$ were scored in each of the four cells of \cref{tab:criteria}. HumanEval+'s optimal
plateau contains the training-split ratio in 4 of 4 cells, $\lambda{=}1$ lies strictly outside it every
time, and $\lambda{=}0$, which ignores false alarms altogether, is never strictly better.

\section{Peer-dependent verdicts}
\label{app:peers}

A level-2 operator reads \op{ctx}, which is built from every raw sample for the task, so its verdict is a
function of that peer electorate and not of the candidate alone. The two effects were measured apart.
Against a frozen electorate, 0 of 35 operators change a single verdict, held-out or training, an exact
identity we check rather than assume, and one every number here relies on. Against an electorate grown
from 3424 to 12744 samples, all 35 change at least one, 1996 flags added and 440 removed, one operator
more than doubling its own flag count on rows whose code never changed. So the wider interface buys
discriminating power at the cost of referential stability, and a deployed level-2 operator has to pin its
electorate the way we pin ours.

The direction of the change is what answers a conformity objection, and it is favourable. Every
frozen selection gains $+0.0009$ to $+0.0036$ under the grown electorate with no material reordering, so
the operators are reading signal out of peer context rather than tracking consensus for its own sake. On
the same draw the five-operator rule of \cref{sec:selection}, selected on the training split and never
re-selected, scores $+0.0220$ ($z{=}8.34$, $p{=}0.0010$, $+79/{-}12$ tasks) on 9320 candidates from three
generators it was never fitted to. Raw deltas are not comparable across the two draws, because the deeper
one is roomier (headroom $0.0995$ against $0.0424$), which is why both are quoted as a fraction of their
own headroom: $22.1\%$ against $18.9\%$. The operators are out of sample there; the task list is not.

\section{Redundancy and the diversity veto}
\label{app:redundancy}

\begin{figure}[t]
\centering
\includegraphics[width=\linewidth]{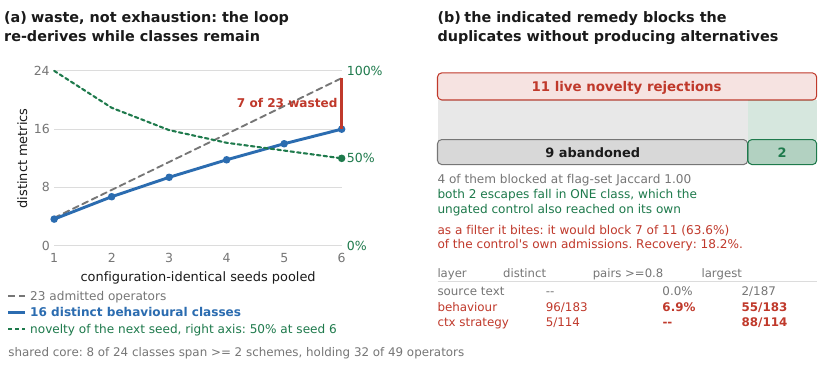}
\caption{\textbf{(a)} Exact rarefaction over all 63 seed subsets and 720 arrival orders, not a
bootstrap: the 23 operators admitted across those six seeds fall into 16 distinct behavioural classes, and the curve does not
saturate. \textbf{(b)} The indicated remedy, run live: what became of each of the 11 candidates a
behavioural-novelty term at admission blocks. The strip below reads redundancy at three layers, textual,
behavioural and strategic, giving for each the distinct classes, the share of pairs at Jaccard
${\geq}0.8$ and the largest class.}
\label{fig:neg}
\end{figure}

\Cref{fig:neg} holds the redundancy evidence behind \cref{sec:limits}. The veto itself was run, not
merely proposed: under the fold rotation of \cref{app:budget} it is at or below the plain union
($+0.0065/{+}0.0057/{+}0.0047$ at $m{=}2,3,4$, harming $9/11/11$ tasks against the union's 8), and
development folds disable it in 5 of 9 directions. On the rediscovery-heavy pool of \cref{sec:limits}
the frozen rule holds on the 5 novel operators ($+0.0040$). It buys parsimony, 7 operators for the
effect of 26, rather than quality. It vetoes \emph{flag-set} overlap at admission, which makes it a weaker
instrument than the behavioural-novelty gate of \cref{sec:limits}, and the two fail differently.

\section{The judge and induction baselines}
\label{app:judge}

\paragraph{The judge, in four configurations.}
The judge sees the task statement, the prompt-visible asserts and the candidate source, which is
exactly the level-1 interface \op{op(task, code)}, and never the hidden suite. Only the 1762 distinct
(task, candidate) pairs that pass the visible checks are judged, because those are the only ones the
endpoint scores. The configurations cross two models, the one that authors operators and the cheaper
Claude Haiku 4.5, which is also one of the candidate generators of \cref{sec:setup}, so the cheap judge
scores some of its own output; and two prompts: a one-word verdict, and a reason-then-verdict form whose wording states the training
split's own class balance, so the judge is given no statistic the operators are denied. Replies are
cached by (pool, task, source hash) and a reply naming no verdict flags nothing, which is the
conservative reading; rescoring the caches twice reproduces byte-identical output.

\begin{table}[h]
\caption{The judge as a metric, on the endpoint and null of \cref{tab:headline}. Precision, recall and
accuracy read the judge as a \emph{classifier} of the same pairs, which is a different question from
the endpoint delta. The last two rows are the same two references as \cref{tab:headline}.}
\label{tab:judge}
\centering\small
\setlength{\tabcolsep}{4.2pt}
\begin{tabular}{@{}llrrrrrrr@{}}
\toprule
model & prompt & $\dlt$ & $p$ & $+/-$ & \% headroom & precision & recall & accuracy \\
\midrule
cheap  & reason-then-verdict & $+0.0070$ & $0.0010$ & $+21/{-}8$ & $16.5$ & $0.403$ & $0.161$ & $0.818$ \\
cheap  & one word            & $+0.0044$ & $0.0130$ & $+20/{-}9$ & $10.3$ & $0.369$ & $0.199$ & $0.807$ \\
author & reason-then-verdict & $+0.0040$ & $0.0040$ & $+14/{-}8$ & \phantom{0}$9.5$ & $0.677$ & $0.154$ & $0.845$ \\
author & one word            & $+0.0036$ & $0.0060$ & $+17/{-}7$ & \phantom{0}$8.6$ & $0.600$ & $0.203$ & $0.844$ \\
\midrule
\multicolumn{2}{@{}l}{hand-written comparator}  & $+0.0069$ & $0.0130$ & $+22/{-}10$ & $16.3$ & & & \\
\multicolumn{2}{@{}l}{loop-authored operator}   & $+0.0065$ & $0.0010$ & $+20/{-}6$ & $15.4$ & & & \\
\bottomrule
\end{tabular}
\end{table}

In \cref{tab:judge}, reasoning before answering raises the cheap model's delta and lowers its flag
count, and both author-model configurations sit at roughly half that delta. As a classifier the author model is the more precise
($0.677$ against $0.403$) and the less useful, which is the recall dominance of \cref{sec:setup}
again: a precise judge that says nothing on most bad candidates cannot change what is left to draw
from. No reply in either reason-then-verdict configuration failed to name a verdict; the one-word
configurations lost 1 of 1762 and 5 of 1762, and an unnamed verdict flags nothing.

\paragraph{Overlap and cost.}
\cref{tab:overlap} scores the two flag sets and their Boolean combinations through the same code. Read
the overlap before the union delta: a union that beats both parts is only interesting when the parts
disagree, and here they do.

\begin{table}[h]
\caption{The loop-authored operator and the best judge configuration as flag sets, on the same endpoint.
Jaccard between the two is $0.105$: 27 flags in common of 257.}
\label{tab:overlap}
\centering\small
\begin{tabular}{@{}lrrrr@{}}
\toprule
flag set & $\dlt$ & $p$ & $+/-$ & flags \\
\midrule
operator only (set difference) & $+0.0010$ & $0.0679$ & $+8/{-}6$ & \phantom{0}99 \\
judge only (set difference)    & $-0.0007$ & $0.6563$ & $+9/{-}9$ & 131 \\
both agree (intersection)      & $+0.0053$ & $0.0010$ & $\mathbf{+13/{-}1}$ & \phantom{0}27 \\
either (union)                 & $\mathbf{+0.0073}$ & $0.0010$ & $+27/{-}12$ & 257 \\
operator                       & $+0.0065$ & $0.0010$ & $+20/{-}6$ & 126 \\
judge                          & $+0.0070$ & $0.0010$ & $+21/{-}8$ & 158 \\
\bottomrule
\end{tabular}
\end{table}

Cost comes from the call ledgers rather than an estimate. The run that wrote the operator made 47
calls, and every candidate it has scored since cost nothing; all eight runs together made 376. The
judge costs one call per candidate with nothing amortised, so a single pass over this endpoint is 1762
calls, $37.5\times$ the run that produced the operator or $4.7\times$ all eight runs including the two
that admitted nothing, and every future candidate costs that again.

\paragraph{Verifier induction.}
Same author model, same system prompts, same four screens in the same order, same training split,
endpoint and null. Two things change deliberately: candidates are commissioned against a balanced
random labelled sample rather than against a signature collision, and they are admitted for predicting
the \emph{label}, at least 3 true positives at precision at least $0.5$, which are the label-space
analogues of the deployed gate's three helped tasks and helped $>$ hurt. \cref{tab:induce} reports
both interfaces under three ways of combining whatever was admitted.

\begin{table}[h]
\caption{Verifier induction at each interface. Combiners are the plain union of admitted checks, a
$\geq2$-of-$k$ vote, and the single check with the best \emph{training} $F_1$; only the last is a rule a
practitioner could apply without held-out labels.}
\label{tab:induce}
\centering\small
\begin{tabular}{@{}lrr@{}}
\toprule
 & level 1, \op{detect(task, code)} & level 2, \op{detect(task, code, ctx)} \\
\midrule
candidates commissioned       & 2 runs $\times$ 24 & 1 run $\times$ 16 \\
admitted                      & 2 & 6 \\
union of the admitted checks  & $+0.0016$ ($p{=}0.19$) & $+0.0053$ ($p{=}0.0120$) \\
$\geq2$-of-$k$ vote            & $+0.0000$ ($p{=}0.52$) & $+0.0053$ ($p{=}0.0010$) \\
best single by training $F_1$ & $+0.0016$ ($p{=}0.19$) & $+0.0054$ ($p{=}0.0060$) \\
\bottomrule
\end{tabular}
\end{table}

Level 1 fails the way our own level-1 search fails: 48 calls, 2 admissions, one run admitting nothing,
and no combination separating from its null. Level 2 works. The screens fired on this baseline as they
do on our own author path, rejecting candidates for source-pattern fitting and for prompt dispatch,
so it was given no exemption our method does not have. Its best \emph{single} check reaches $+0.0074$,
above both the comparator and our operator, but that check is chosen by reading held-out deltas and is
not a legitimate headline. The legitimate reading is the warning: on this endpoint the ceiling
reachable by predictive-fit admission at the wider interface is at least as high as the one we report.

\paragraph{Label prediction as a composition criterion.}
The same criterion swap can be made with no model and no sampling at all, because
\cref{sec:selection}'s enumeration puts every subset in memory, so a criterion is a total order over a
finite set and its choice, delta and percentile are counts rather than estimates. Carrying $J$ as the
instrument check, its argmax reproduces the published one in all four cells.

\begin{table}[h]
\caption{Held-out $\dlt$ and exact landscape percentile of four criteria across the four cells of
\cref{tab:criteria}. C8 is precision under the same floor of 3 true positives, and it is degenerate with
the floor in place: it picks the same two-operator set flagging ten tasks in every cell.}
\label{tab:critfit}
\centering\small
\setlength{\tabcolsep}{4.2pt}
\begin{tabular}{@{}lcccc@{}}
\toprule
criterion & 26 ops, full & 35 ops, full & 26 ops, frozen & 35 ops, frozen \\
\midrule
C1, $J = \mathrm{TP}-\mathrm{FP}$   & $+0.0155$ (56.3) & $+0.0140$ (42.3) & $+0.0220$ (85.7) & $+0.0079$ \phantom{0}(4.9) \\
C6, balanced accuracy               & $\mathbf{+0.0286}$ (98.7) & $\mathbf{+0.0253}$ (91.1) & $\mathbf{+0.0236}$ (92.7) & $\mathbf{+0.0236}$ (94.1) \\
C7, label-prediction $F_1$          & $+0.0244$ (89.2) & $+0.0231$ (85.7) & $+0.0219$ (85.0) & $+0.0218$ (85.8) \\
C8, precision                       & $+0.0038$ \phantom{0}(0.0) & $+0.0038$ \phantom{0}(0.0) & $+0.0038$ \phantom{0}(0.3) & $+0.0038$ \phantom{0}(0.1) \\
\bottomrule
\end{tabular}
\end{table}

\cref{tab:critfit} adds two criteria to the six of \cref{tab:criteria}, and one of them wins against
$J$: C7 beats it in three cells, matches it in the fourth ($+0.0219$ against $+0.0220$, inside the
$0.002$ tolerance the study calls a tie), and does not collapse when the pool grows from 26 operators
to 35, which $J$ does. So the gap
\cref{sec:selection} measures is partly recoverable by the criterion the related work already uses, and
C6 recovers more of it in all four cells. C7 selects a different subset from $J$ each time, at Jaccard
$0.22$ to $0.67$ against $J$'s choice, so this is not two names for one ranking.

\section{Broader impacts}

Used as intended, the loop makes a metric's validity a measured quantity rather than an assumed one, and
what it ships costs nothing per candidate to run, so the positive case is a cheap screen that catches
faulty answers before a person or a downstream agent accepts one. The negative case runs through the same
mechanism: an operator is admitted on its measured effect over a sample, so a pool that helps where it was
gated can be silently blind elsewhere, and a rule that flags nothing on some subpopulation changes what
survives for that subpopulation alone. Reporting the flag budget, the headroom and the electorate a rule
depends on, as \cref{sec:payoff} and \cref{sec:limits} do, is what keeps that visible rather than assumed,
and it is why we confine the loop here to a setting where an exact oracle can check the claim.

\end{document}